\documentclass[10pt,twocolumn,letterpaper]{article}

\usepackage{cvpr}              
\definecolor{cvprblue}{rgb}{0.21,0.49,0.74}
\usepackage[pagebackref,breaklinks,colorlinks,allcolors=cvprblue]{hyperref}
\usepackage{multirow}
\usepackage{amsmath}
\usepackage{amsthm}
\newtheorem{lemma}{Lemma}
\newtheorem{theorem}{Theorem}
\usepackage{algorithm}
\usepackage{algpseudocode}
\usepackage{dsfont} 
\usepackage{xcolor}
\usepackage{pifont}
\newcommand{\cmark}{\ding{51}}  
\newcommand{\xmark}{\ding{55}}  
\usepackage{rotating}
\usepackage{makecell}
\def\paperID{xxxxx} 
\def\confName{CVPR}
\def\confYear{2026}

\title{Foundation Model Priors Enhance Object Focus in Feature Space for Source-Free Object Detection}

\author{
Sairam VCR$^{1}$ \quad
Rishabh Lalla$^{2}$ \quad
Aveen Dayal$^{1}$ \quad
Tejal Kulkarni$^{3}$ \\
Anuj Lalla$^{4}$ \quad
Vineeth N.\ Balasubramanian$^{1,5}$ \quad
Muhammad Haris Khan$^{2}$ \\
\\
$^{1}$ IIT Hyderabad \quad
$^{2}$ MBZUAI, Abu Dhabi\quad
$^{3}$ UC San Diego \\
$^{4}$ IIT Jodhpur \quad
$^{5}$ Microsoft Research India
}

\begin{document}
\maketitle
\begin{abstract}
Current state-of-the-art approaches in Source-Free Object Detection (SFOD) typically rely on Mean-Teacher self-labeling. However, domain shift often reduces the detector’s ability to maintain strong object-focused representations, causing high-confidence activations over background clutter. This weak object focus results in unreliable pseudo-labels from the detection head. While prior works mainly refine these pseudo-labels, they overlook the underlying need to strengthen the feature space itself. We propose FALCON-SFOD (Foundation-Aligned Learning with Clutter suppression and Noise robustness), a framework designed to enhance object-focused adaptation under domain shift. It consists of two complementary components. SPAR (\underline{S}patial \underline{P}rior-\underline{A}ware \underline{R}egularization) leverages the generalization strength of vision foundation models to regularize the detector’s feature space. Using class-agnostic binary masks derived from OV-SAM, SPAR promotes structured and foreground-focused activations by guiding the network toward object regions. IRPL (\underline{I}mbalance-aware Noise \underline{R}obust \underline{P}seudo-\underline{L}abeling) complements SPAR by promoting balanced and noise-tolerant learning under severe foreground-background imbalance. Guided by a theoretical analysis that connects these designs to tighter localization and classification error bounds, FALCON-SFOD achieves competitive performance across SFOD benchmarks.
\end{abstract}

\section{Introduction}
\label{sec:intro}

\begin{figure*}[t]
    \centering
\includegraphics[width=\linewidth]{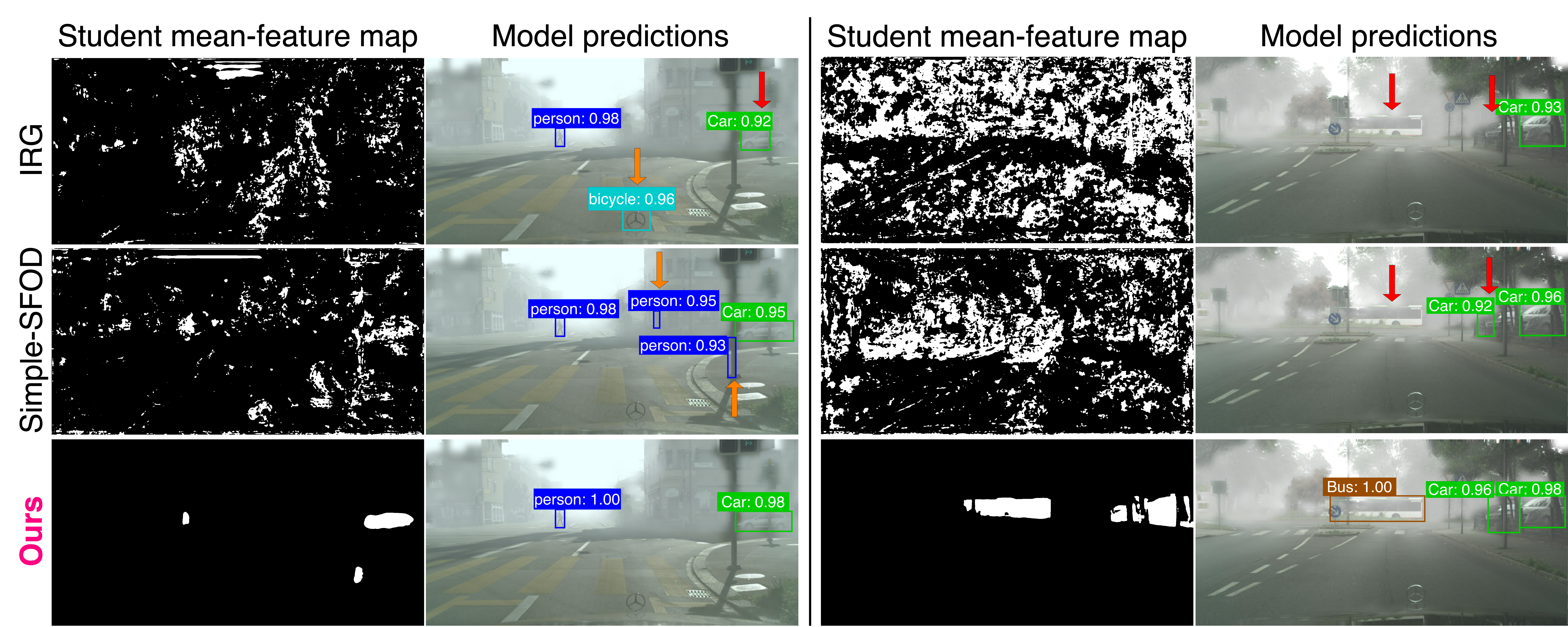}
    \vspace{-12pt}
    \caption{\textbf{Towards Object-Focused Representations in SFOD.} Two examples from the Foggy Cityscapes~\citep{sakaridis2018semantic} target set. In each example, the two columns show the student’s thresholded channel-mean feature maps from the last backbone layer (brighter = higher activation) and the model predictions, respectively. Existing state-of-the-art methods IRG~\citep{irg} and Simple-SFOD~\citep{hao2024simplifying} tend to produce less object-focused activations extending into background regions, leading to less precise object localization (\textcolor{red}{red arrows}) and false positives (\textcolor{orange}{orange arrows}). In contrast, our method produces compact, object-shaped activations with tighter boxes and accurate labels, demonstrating stronger spatial coherence and object awareness. \emph{(Best viewed when zoomed in)}}
    \label{fig:teaser}
    \vspace{-10pt}
\end{figure*}

Source-Free Object Detection (SFOD) aims to adapt a detector trained on labeled source data to an unlabeled target domain without accessing any source samples during adaptation. This setting is practically important, as it removes the need to store or share potentially sensitive source data, a common constraint in real-world applications such as autonomous driving, surveillance, and medical imaging. Current state-of-the-art SFOD approaches~\cite{irg, pets, hao2024simplifying, khanh2024dynamic} largely adopt the \textit{Mean Teacher} self-labeling framework. In this paradigm, a teacher network, maintained as an exponential moving average (EMA) of the student, generates pseudo-labels on target images to facilitate adaptation. Although this framework effectively exploits unlabeled data, it inherits a strong source-domain bias. Since the teacher is trained exclusively on source data, it often produces unreliable pseudo-labels when applied to the target-domain~\cite{sfod, sfod_mosaic, a2sfod, lods, irg, pets, hao2024simplifying, khanh2024dynamic}. These noisy labels are subsequently propagated to the student, thereby compromising adaptation stability. To alleviate this issue, existing methods refine pseudo-label selection~\cite{sfod, sfod_mosaic, pets, khanh2024dynamic}, exploit object relationships within the target-domain~\cite{irg}, enforce style invariance~\cite{lods}, or apply adversarial alignment strategies~\cite{a2sfod}. However, these approaches largely overlook a more fundamental limitation. We observe that \emph{domain shift weakens the detector’s \textbf{object focus}}, wherein feature activations become spatially diffuse and extend into background clutter, reducing the clarity of object boundaries, as demonstrated in Figure \ref{fig:teaser}. As a result, the detection head operates on less discriminative features, producing suboptimal pseudo-labels. This observation motivates us to strengthen SFOD from the feature level by promoting structured, object-centered representations rather than solely refining pseudo-labels.

Recent progress in cross-domain object detection~\cite{lavoie2025large, han2025vfm} demonstrates that large-scale vision foundation models can serve as reliable, domain-agnostic knowledge sources. Motivated by these advances, we explore their potential in a source-free setting and propose \textbf{SPAR} (\underline{S}patial \underline{P}rior-\underline{A}ware \underline{R}egularization), a framework that uses class-agnostic binary masks derived from a frozen segmentation model~\cite{yuan2024open} to regularize the detector’s feature space. 
SPAR encourages the student’s channel-mean activations to align with these masks, enhancing foreground structure and promoting object-focused responses. While SPAR improves the spatial quality of learned features, pseudo-labels in detection remain highly imbalanced, with abundant background regions and scarce positive samples. To stabilize training under this imbalance and mitigate residual class-label noise, we introduce \textbf{IRPL} (\underline{I}mbalance-aware Noise \underline{R}obust \underline{P}seudo-\underline{L}abeling). It is a classification loss designed to promote balanced and noise-tolerant learning by down-weighting overconfident teacher-student agreements, reinforcing underrepresented categories, and countering the inherent background dominance in detection. Together, SPAR and IRPL complement each other: SPAR strengthens the structural integrity of object representations, while IRPL ensures balanced and robust supervision. We further provide a theoretical analysis linking these contributions to tighter localization and classification error bounds. Overall, our framework advances object-focused adaptation under domain shift. Our main contributions are summarized as follows:

\begin{itemize}
    \item To the best of our knowledge, this is the first effort to identify and demonstrate the importance of \textbf{object-focused feature representation} in advancing SFOD.
    \item We propose \textbf{SPAR}, a spatial prior-based regularization that promotes structured, object-centered features, and \textbf{IRPL}, an imbalance-aware noise robust pseudo-labeling loss for stable adaptation under noisy supervision.
    \item We provide one of the first \textbf{theoretical risk-bound analysis} for SFOD, linking our losses to tighter error bounds.
    \item Our approach achieves competitive performance across standard SFOD benchmarks. 
\end{itemize}

\section{Related Works}
\label{sec:related_works}
\paragraph{Unsupervised Domain-Adaptive Object Detection (UDAOD).}
UDAOD adapts a source-trained detector to an unlabeled target-domain when source data is available during adaptation~\citep{robust, mega_cdn, munir2021ssal, kennerley2024cat, li2025seen}. 
Recent works like DINO Teacher~\citep{lavoie2025large} leverage foundation models by training a source-domain labeller with a frozen DINOv2 and then aligning the student’s patch features to a frozen DINO encoder. While this approach enriches supervision, it remains computationally demanding as the DINO encoder is used in an online manner for target adaptation. Our \textbf{SPAR} loss provides a contrasting design: it uses class-agnostic binary masks from a frozen open-vocabulary segmenter~\cite{yuan2024open} that are computed once before adaptation, so the foundation model is not queried during training or inference and the added cost is a one-time preprocessing step.

\vspace{-12pt}
\paragraph{Source-Free Object Detection (SFOD).}
SFOD eliminates the need to access source data during adaptation~\citep{irg, pets, hao2024simplifying, varailhon2024source, khanh2024dynamic}. IRG~\citep{irg} refines pseudo-labels via instance-relation graphs, PETS~\citep{pets} stabilizes teacher-student training with periodic exchanges, and \emph{Simple-SFOD}~\citep{hao2024simplifying} shows that careful self-training design can outperform complex architectures. SF-YOLO~\citep{varailhon2024source} adapts the YOLO detector in a source-free manner using target domain-specific augmentations and a teacher-student communication mechanism. DRU~\citep{khanh2024dynamic} dynamically retrains and updates the mean teacher to enhance stability and performance during adaptation. While these works focus on refining pseudo-labels or stabilizing training, they overlook the importance of object-focused feature representation. Our proposed model addresses both the problems in a lightweight (simple-by-design, zero inference overhead), effective manner. 

\vspace{-8pt}
\paragraph{Noise-Robust Learning with Noisy Labels.}
Noisy-label learning introduces objectives that suppress the impact of corrupted labels~\citep{wang2024epsilon}. 
However, a direct adaptation of ~\citep{wang2024epsilon} fails in SFOD (detailed in Section \ref{sec:ablations}) due to proposal imbalance, heavy class skew, and confirmation bias in teacher-student agreements. We extend it with foreground/background weighting, and entropy regularization, which together make IRPL effective for detection.
\section{Methodology}
\label{sec:method}

\subsection{Preliminaries}
\label{sec:prelims}
\vspace{-2pt}
\textbf{Problem Statement.} We denote the labeled source-domain dataset as $ \mathcal{D}_S=\{(x_i^{s},\,\mathcal{Y}_i^s)\}_{i=1}^{N_S} $, where $ x_i^s $ denotes the $ i^{\text{th}} $ source image and $ \mathcal{Y}_i^s $ is the corresponding ground-truth annotation containing bounding-box locations and class labels.  
$ \mathcal{Y}_i^s=\{(b_{ij},c_{ij})\}_{j=1}^{O_i} $, where $ b_{ij}\in\mathbb{R}^4 $ denotes bounding-box coordinates, $ c_{ij}\in\{0,\dots,K\} $ denotes the class label for the $ j^{\text{th}} $ object, and $ O_i $ denotes the number of objects in image $ x_i^s $.  
The unlabeled target-domain dataset is denoted by $ \mathcal{D}_T=\{x_i^{t}\}_{i=1}^{N_T} $.  
The cardinalities of source and target-domain images are denoted by $ N_S $ and $ N_T $ respectively.  
An object detector can be expressed as $h(x)=f\!\bigl(g(x)\bigr)$,
where \(g\) is the feature extractor and \(f=(f_c,f_r)\) contains the: (i) \textit{Classification head}: \(f_c\!\bigl(g(x)\bigr)\in\Delta^{K}\)
        (a softmax over \(K+1\) classes, with the extra class for background); and (ii) \textit{Regression head}: \(f_r\!\bigl(g(x)\bigr)\in\mathbb{R}^{4}\),
        predicting the bounding-box coordinates.
The detector is usually trained by minimizing the combination of classification and regression loss terms:
$\mathcal{L}
    =\mathcal{L}_{\mathrm{cls}}
    +\mathcal{L}_{\mathrm{reg}}$.
Most existing efforts employ cross-entropy (or a variant) for the
classification term and an \(L_{1}\)-style loss for the regression term.
The goal of \textbf{SFOD} is to adapt a detector $h_{pre}$ trained on labeled
source data \(\mathcal{D}_S\) to unlabeled target data \(\mathcal{D}_T\),
without any further access to \(\mathcal{D}_S\).

\vspace{2pt}
\noindent\textbf{Mean-Teacher based Self-Training Framework.}\; Current state-of-the-art SFOD methods adopt a \emph{teacher-student} strategy that follows the Mean-Teacher (MT) paradigm.
A \emph{student} detector $h^{st}$ with parameters $\boldsymbol{\Theta}^{st}$ is updated via gradient descent, and a \emph{teacher} detector $h^{te}$ with parameters $\boldsymbol{\Theta}^{te}$ tracks the student through an exponential moving average (EMA).
For every unlabeled target image $x_i^{t}$, a weak augmentation $\tilde{x}_i^{t}$ is first applied and then passed through the teacher to obtain a set of predictions. After standard post-processing operations such as score filtering, non-maximum suppression (NMS), and a confidence threshold, the remaining predictions constitute the pseudo-annotation $\hat{\mathcal{Y}}_i^t = \{(\hat{b}_{ij}, \hat{c}_{ij})\}_{j=1}^{\hat{O_i}}$, where $\hat{b}_{ij}$, $\hat{c}_{ij}$ denote the pseudo bounding boxes and their corresponding pseudo class labels, respectively. $\hat{O_i}$ is the number of pseudo-annotations on the $i^{th}$ image.
The student is then trained on a \emph{strongly} augmented view $\bar{x}_i^{t}$ by minimizing the sum of the classification and localization losses of the detector with respect to these pseudo labels:
\vspace{-1pt}
\begin{equation}
\mathcal{L}_{\mathrm{MT}}
=\mathcal{L}_{\mathrm{cls}}\bigl(\bar{x}_i^{t},\hat{\mathcal{Y}}_i^t\bigr)
+\mathcal{L}_{\mathrm{loc}}\bigl(\bar{x}_i^{t},\hat{\mathcal{Y}}_i^t\bigr).
\label{eq:mt-loss}
\end{equation}
The network parameters are updated by:
\vspace{-1pt}
\begin{equation}
\begin{aligned}
\boldsymbol{\Theta}^{st} &\leftarrow \boldsymbol{\Theta}^{st} - \eta\,\nabla_{\boldsymbol{\Theta}^{st}}\mathcal{L}_{\mathrm{MT}} \\[2pt]
\boldsymbol{\Theta}^{te} &\leftarrow \delta\,\boldsymbol{\Theta}^{te} + (1-\delta)\,\boldsymbol{\Theta}^{st}
\end{aligned}
\label{eq:update}
\end{equation}
where $\eta$ is learning rate and $\delta \!\in\!(0,1)$ is EMA decay factor.  

\vspace{-8pt}
\paragraph{Theory-guided Objective (forward reference to Sec.~\ref{sec:theory}).}
The detection-risk decomposition in Sec.~\ref{sec:theory} will show that training on noisy pseudo-labels inflates (i) the \emph{classification} risk by a factor $1/\lambda$ (Lemma~\ref{lem:ce-general}, Theorem~\ref{thm:det-bound-general}) and (ii) the \emph{localization} risk through two additive terms, the deviation $\eta_{\text{reg}}$ and the miss-rate $\zeta$ (Lemma~\ref{lem:reg-missing}, Theorem~\ref{thm:det-bound-general}). We therefore design two complementary modules: a robust classification loss (\textbf{IRPL}) that replaces the multiplicative inflation with a tighter additive term (Theorem~\ref{thm:epsilon-excess}), and a spatial-focus regularizer (\textbf{SPAR}) that directly shrinks $\eta_{\text{reg}}$ and $\zeta$ by mitigating feature-space misalignment.

\subsection{FALCON-SFOD}
\label{sec:falcon-sfod}
\vspace{-2pt}
\textbf{Motivation and Overview.}
As observed in \citep{irg,pets,hao2024simplifying, khanh2024dynamic}, a prevalent problem in mean-teacher self-training is that pseudo class-labels tend to be noisy due to domain shifts. \emph{Beyond this well-known issue, we identify that domain shift weakens the detector’s object-focus} in SFOD: spatial activations for true objects are diluted by background clutter, which degrades localization and cascades into misclassification (see Fig.~\ref{fig:teaser}). 
We propose \textbf{FALCON-SFOD}, coupling two dedicated objectives within the standard Mean-Teacher framework: \textbf{SPAR} for spatial focus regularization and \textbf{IRPL} for pseudo class-label robustness. These losses are integrated with the conventional localization objective to enhance robustness during source-free adaptation. 

\noindent \textbf{Spatial Prior-Aware Regularization (SPAR).}
Consider the detector \(h(x)=f(g(x))\) (Sec.~\ref{sec:prelims}).  
The feature extractor outputs an activation tensor \(a=g(x)\in\mathbb{R}^{H\times W\times C}\), where \(H\), \(W\), and \(C\) are its height, width, and channel count.  
Taking the channel-wise mean yields \(A\in\mathbb{R}^{H\times W}\), which highlights spatial locations with high average activations and thus usually traces foreground objects~\cite{rebbapragada2024c2fdrone}.
Under a domain shift, however, this map becomes contaminated by background clutter:
true objects may fade while background areas are spuriously accentuated
(as illustrated in Fig.~\ref{fig:teaser}).  We derive a \emph{foreground prior} once, offline, by running a frozen open-vocabulary segmentation model on target images. We discard the class information and binarize these masks, where pixels belonging to any segmented region are set to \(1\) (foreground) and the rest to \(0\) (background). The prior is not queried during training or inference. For a target image $x_i^t$, let $A_G(x_i^t)\in[0,1]^{H'\times W'}$ be the prior mask and $A_S(x_i^t)\in[0,1]^{H\times W}$ be the student channel-mean map rescaled to $H'\times W'$ to match the prior. We encourage agreement using a mean $\ell_1$ term plus Dice::
\begin{align}
\mathcal{L}_{\mathrm{SPAR}}(x_i^t)
&= \frac{\lambda_1}{H'W'} \sum_{j,k} \lvert A_{\mathrm S}[j,k]-A_{\mathrm G}[j,k]\rvert
\notag\\[-2pt]
&\quad +\, \lambda_2\!\left(
1 - \frac{2\sum_{j,k} A_{\mathrm S}[j,k]A_{\mathrm G}[j,k]}
           {\sum_{j,k} A_{\mathrm S}[j,k]+\sum_{j,k} A_{\mathrm G}[j,k]+\varepsilon}
\right)
\label{eq:spar}
\end{align}

where $(j,k)$ index spatial positions, and $\varepsilon$ is a small constant for numerical stability. A hyperparameter sweep in the range (0,5), gave optimal results at $\lambda_1$ = 1 and $\lambda_2$ = 2. More details are presented in \texttt{supplementary}. 

\vspace{4pt}
\noindent \textbf{Imbalance-aware Noise Robust Pseudo-Labeling (IRPL).} Most state-of-the-art SFOD methods minimize cross-entropy loss \citep{irg, pets, hao2024simplifying}, which has an unbounded gradient that allows a single corrupted pseudo class-label to dominate training. Additionally, object detection has an inherent foreground-background imbalance which affects the stability of the student training. 
We therefore introduce IRPL, which mitigates this negative impact of noisy pseudo class-labels by adaptively recalibrating the per-box classification loss and effectively re-weights foreground-vs-background in the loss. 
For student probabilities \(f_c^{st}\!\in\!\mathbb{R}^{K+1}\),  let \(\mathbf{p}=f_c^{st}\) and
$ t = \arg\max_{k} p_k $. The following transform rescales the peak while making sure that elements of $\mathbf{p}'$ sum up to 1.
\begin{equation}
p'_{k}=
\begin{cases}
\dfrac{p_k+m}{1+m}, & k=t,\\[4pt]
\dfrac{p_k}{1+m},   & k\neq t,
\end{cases}
\label{eq:eps_softmax}
\end{equation}
where $m$ is a large real value.

For each image $x_i^{t}$, we minimize
\begin{equation}
\begin{split}
\mathcal{L}_{\mathrm{IRPL}}
  =&\!\!\sum_{(\hat{b},\hat{c})\in\hat{\mathcal{Y}}_i^t}
        w_{\hat{c}}\!\left[\alpha(-\!\log p'_{\hat{c}})
        +\beta(1-p_{\hat{c}})\right] \\
   &+\gamma\,D_{\mathrm{KL}}\!\bigl(
       \bar{\mathbf p}\,\big\|\,\mathcal{U}_{\mathcal{K}}
     \bigr).
\end{split}
\label{eq:irpl}
\end{equation}

where 
\begin{equation}
\begin{split}
\mathcal{K}&:=\{0,\dots,K\!-\!1\}, \quad
Z=\!\!\sum_{(\hat{b},\hat{c})\in\hat{\mathcal{Y}}_i^t}\sum_{k\in\mathcal{K}}p_{k};\\[3pt]
\bar p_{k}&=\frac{1}{Z}\!\sum_{(\hat{b},\hat{c})\in\hat{\mathcal{Y}}_i^t}\!p_{k},
\quad \forall k\!\in\!\mathcal{K};\\[3pt]
D_{\mathrm{KL}}\!\bigl(\bar{\mathbf p}\,\|\,\mathcal{U}_{\mathcal{K}}\bigr)
&=\log|\mathcal{K}|+\!\sum_{k\in\mathcal{K}}\bar p_{k}\log\bar p_{k}.
\end{split}
\label{eq:definitions}
\end{equation}

\noindent We provide detailed hyperparameter analysis in the \texttt{supplementary}.
\noindent \paragraph{Why IRPL is intrinsically robust to noisy pseudo-labels?}
The \emph{peak–adjust} operation in Eq.~\ref{eq:eps_softmax} moderates the student’s logits by adding a large margin $m$ to its highest probability and then renormalizing.  
This creates two mutually exclusive regimes:  1) \textbf{Teacher and student agree} ($\hat c = t$).  
        The margin sits on the \emph{same} logit that the loss differentiates, so the cross-entropy gradient for that box is uniformly scaled by the factor $p_{\hat c}/(p_{\hat c}+m) \ll 1$.  
        Easy, likely-clean boxes therefore contribute vanishing updates, acting as a built-in \emph{soft early-stopping} mechanism that prevents over-fitting to already-correct labels. 2) \textbf{Teacher and student disagree} ($\hat c \neq t$).  
        The margin affects a \emph{different} logit; the derivative with respect to the true class is unchanged, and the gradient reduces to the standard cross-entropy form.  
        Hard or potentially mislabeled boxes retain a full corrective signal, allowing the student to challenge erroneous teacher guidance.

\section{Theoretical Insights}
\label{sec:theory}
This section formalizes how the modules introduced in Sec.~\ref{sec:method} target specific terms in the detection risk. We first decompose risk under teacher-generated pseudo-labels, then show that a peak-adjusted classification objective yields a tighter classification term and explain how spatial confusion appears additively in the localization term, precisely what SPAR is designed to reduce.

For a given dataset $\mathcal{D} = \{(x_i,y_i)\}_{i=1}^{N}$, we define an object set for image $i$ as $\mathcal{Y}_i = \{ (b_{ij},c_{ij})\}_{j=1}^{O_i}$, the pseudo annotations as $\hat{\mathcal{Y}_i} = \{ (\hat{b_{ij}},\hat{c_{ij}})\}_{j=1}^{\hat{O_i}}$ and the detection risk is shown in Eq. \ref{det_risk}. We implicitly index over objects inside each image.  We write $f(g(x))$ simply as $f(x)$ for brevity, $f_c\colon x\mapsto(p_0(x),\dots,p_K(x))$, and $\mathcal{D}_{X,C}$ and $\mathcal{D}_{X,B}$ as the marginal of $\mathcal{D}$ over the image-class $(x,c)$ pairs and image-box pairs $(x,b)$, respectively.
\begin{equation}
\label{det_risk}
\begin{split}
R_{\mathcal{D}}^{det}(f)
&= R^{cls}_{\mathcal{D},clean}(f_c)
  + R^{reg}_{\mathcal{D},clean}(f_r) \\
&= \mathbb{E}_{(x,c)\sim \mathcal{D}_{X,C}}
    [-\log p_c(x)] \\
&\quad
  + \mathbb{E}_{(x,b)\sim \mathcal{D}_{X,B}}
    \bigl\|f_r(x)-b\bigr\|_1
\end{split}
\end{equation}
where $R^{cls}_{\mathcal{D},clean}$ is the standard cross entropy loss and $R^{reg}_{\mathcal{D},clean}$ is the standard $L_1$ regression loss used in most of the object detectors. We now state in Lemma \ref{lem:ce-general} the classification risk under noisy pseudo class-labels.
\begin{lemma}
\label{lem:ce-general}
Let $\mathcal{D}^T$ be the target distribution over $(x,c)$, and let the
pseudo-label $\hat c$ be drawn from an \emph{arbitrary}
class-conditional transition matrix
$T\in[0,1]^{(K+1)\times (K+1)}$ with
$\sum_{i=0}^{K}T_{ji}=1$ for every $j$ and $\lambda\;=\;\min_{j}T_{jj}\;>\;0.$
For any classifier $f^{st}_c:x\mapsto p(x)=\bigl(p_0(x),\dots,p_K(x)\bigr)$
and $ R^{cls}_{\mathcal{D}^T,\;clean}(f^{st}_c)
  \;=\;
  \mathbb{E}_{(x,c)\sim\mathcal{D}^T}
        \bigl[-\log p_c(x)\bigr]$, $
  R^{cls}_{\mathcal{D}^T,\;noisy}(f^{st}_c)
  \;=\;
  \mathbb{E}_{(x,\hat c)}
        \bigl[-\log p_{\hat c}(x)\bigr],$ we have the following relationship
\begin{equation}
  R^{cls}_{\mathcal{D}^T,\;clean}(f^{st}_c)
  \;\le\;
  \frac{1}{\lambda}\;
  R^{cls}_{\mathcal{D}^T,\;noisy}(f^{st}_c).
  \label{eq:general-noise-bound}
\end{equation}
\end{lemma}

Lemma~\ref{lem:ce-general} employs the class-conditional transition matrix $T\!\in\![0,1]^{(K+1)\times (K+1)}$, where  $T_{ji}= \Pr[\hat c=i \mid c=j]$ represents the label noise induced by the mean-teacher: an exponential-moving-average (EMA) copy of the student whose deterministic predictions form a stochastic channel over the data
distribution \citep{mean_teacher}. The diagonal element $T_{jj}$ is the teacher’s per-class hit-rate, and we define
$\lambda=\min_j T_{jj}>0$ to rule out classes the teacher never recognizes, a minimal condition for identifiability of the clean risk under arbitrary label noise \citep{liu2023identifiability}. Lemma~\ref{lem:ce-general} therefore shows that mean-teacher asymmetry costs only a multiplicative factor $1/\lambda$; when the teacher is perfect ($\lambda=1$) the bound reduces to the standard clean-risk expression. Next, we show in Lemma \ref{lem:reg-missing} the regression risk under noisy pseudo bounding box labels.

\begin{lemma}
\label{lem:reg-missing}
Let $\mathcal{D}^T$ be the target-domain distribution over image–box
pairs $(x,b)$ and let $f^{te}_r$ be the teacher regressor that outputs a
pseudo-box $\hat b=f^{te}_r(x)$.
For every ground-truth box define the indicator
  $M(x,b)\;=\;
  \mathds 1\!\bigl[\;
    \mathrm{IoU}\bigl(\hat b,b\bigr)\;\ge\;\tau
  \bigr]\;\in\;\{0,1\},$
i.e.\ $M=1$ when the teacher matches the ground truth box under the usual IoU
threshold~$\tau$, and $M=0$ otherwise.  
Assume all boxes are normalized to the unit square, so that
$\|u-v\|_1\le 2$ for any two boxes $u,v$.
Define
\begin{equation}
\label{eq:reg_noisy}
\begin{split}
R^{reg}_{\mathcal{D}^T,\;noisy}(f^{st}_r)
&= \mathbb{E}_{(x,b)}\!\bigl[
      M\,\|f^{st}_r(x)-\hat b\|_1
    \bigr], \\[4pt]
\eta_{reg}
&= \mathbb{E}_{(x,b)}\!\bigl[
      M\,\|\hat b-b\|_1
    \bigr], \\[4pt]
\zeta
&= \mathbb{E}_{(x,b)}[\,1-M\,].
\end{split}
\end{equation}
Then for any student regressor $f^{st}_r$
\begin{equation}
  R^{reg}_{\mathcal{D}^T,\;clean}(f^{st}_r)
  \;\le\;
  R^{reg}_{\mathcal{D}^T,\;noisy}(f^{st}_r)
  \;+\;
  \eta_{reg}
  \;+\;
  2\,\zeta.
  \label{eq:reg-bound-missing}
\end{equation}
\end{lemma}

Lemma \ref{lem:reg-missing} expresses the clean localization risk as the sum of the noisy risk and the single constant $\eta_{\text{reg}}$, defined as the teacher’s expected $L_{1}$ deviation from ground truth and entirely determined by the geometry of boxes; no distributional assumptions are introduced. Because the argument is purely metric, the bound holds regardless of how pseudo-boxes are generated or how label noise correlates across objects and classes.  
Importantly for SFOD, domain shift-induced foreground-background misalignment increases both the teacher’s miss-rate $\zeta$ and the deviation term $\eta_{\text{reg}}$ by spreading activations into cluttered background, thereby degrading localization (precisely what SPAR is designed to mitigate in Eq.~\ref{eq:spar}). We now leverage Lemma \ref{lem:ce-general} and Lemma \ref{lem:reg-missing} to state the upper bound on detection risk as seen in Theorem \ref{thm:det-bound-general}.

\begin{theorem}
\label{thm:det-bound-general}
Given pseudo class-labels generated by a teacher with transition matrix $T$ satisfying $\lambda=\min_j T_{jj}>0$ and 
bounding-box pseudo-labels satisfying the noise rate $\eta_{reg}$ and let $\zeta=\mathbb E_{(x,b)}[\,1-M(x,b)\,]$ be the teacher’s miss-rate for ground-truth boxes. Then, for any student heads $(f_c^{st},f_r^{st})$, we have
\begin{equation}
\label{eq:det-bound-general}
\begin{split}
R^{\det}_{\mathcal{D}^T}\!\bigl(f_c^{st},f_r^{st}\bigr)
&\le \frac{1}{\lambda}\,
   R^{cls}_{\mathcal{D}^T,\;noisy}\!\bigl(f_c^{st}\bigr) \\[4pt]
&\quad +\;
   R^{reg}_{\mathcal{D}^T,\;noisy}\!\bigl(f_r^{st}\bigr)
   + \eta_{reg} + 2\,\zeta.
\end{split}
\end{equation}
\end{theorem}

Theorem~\ref{thm:det-bound-general} adds the two sources of error.  
The classification part is inflated by $1/\lambda$, while the regression part is simply shifted by $\eta_{\text{reg}}$ plus the miss-rate penalty $2\zeta$.  
Foreground–background confusion acts precisely through these localization terms, motivating an explicit spatial regularizer that reduces $\eta_{\text{reg}}$ and $\zeta$ by cleaning activations (SPAR, Eq.~\ref{eq:spar}).

\begin{theorem}
\label{thm:epsilon-excess}
Let $f_\eta^*=\arg\min_{f^{st}_c} R^{\eta}_{L}(f^{st}_c)$ be the population
minimizer of the peak-adjusted classification loss
$R^{\eta}_{L}$ under the teacher-noise model $T$. We define $R^{\eta}_{L}(f^{st}_c)\;=\;\mathbb{E}_{(x,c)}
\Bigl[(1-\eta_{x})\,L\!\bigl(f^{st}_c(x),c\bigr)\;+\;
\sum_{k\neq c}\eta_{x,k}\,L\!\bigl(f^{st}_c(x),k\bigr)
\Bigr].$ where $L$ is any classification loss that satisfies $\Bigl|\sum_{k=1}^{K}\bigl(L(u_{1},k)-L(u_{2},k)\bigr)\Bigr|
  \;\le\; \delta \text{ whenever } \|u_{1}-u_{2}\|_{2}\le\epsilon,$ 
and $\delta\to 0$ as $\epsilon\to 0$, $\eta_{x,k}\;:=\;\Pr_{T}(\hat c=k\mid x),
\eta_{c}\;:=\sum_{k \neq c}\eta_{x,k}.$ 
  $w=\mathbb{E}_{x}(1-\eta_{c}), 
  a=\min_{x,k}\bigl(1-\eta_{c}-\eta_{x,k}\bigr),$, 
and let $\zeta=\mathbb E_{(x,b)}[\,1-M(x,b)\,]$ be the teacher’s
\emph{miss-rate} for ground-truth boxes, as introduced in
Lemma~\ref{lem:reg-missing}.
Then, for any student regression head $f^{st}_r$,
\begin{equation}
\label{eq:epsilon-excess-det}
\begin{split}
R^{\det}_{\mathcal D^{T}}\!\bigl(f_\eta^*,f^{st}_r\bigr)
&\le
  \Bigl(2\delta+\tfrac{2w\,\delta}{a}\Bigr) \\[4pt]
&\quad
  + R^{\text{reg}}_{\mathcal D^{T},\text{noisy}}\!\bigl(f^{st}_r\bigr)
  + \eta_{\text{reg}}
  + 2\,\zeta.
\end{split}
\end{equation}
\end{theorem}

\noindent
Theorem~\ref{thm:epsilon-excess} replaces the multiplicative factor \(1/\lambda\) in Theorem~\ref{thm:det-bound-general} with the additive term \(2\delta+\tfrac{2w\delta}{a}\), thus making it tighter. Because \(\delta \!\to\! 0\) as \(\varepsilon \!\to\! 0\), this additive bound becomes arbitrarily tight even for moderate \(\lambda\); moreover, since \(1/\lambda \ge 1\), it is \emph{strictly tighter} than the original multiplicative bound whenever the teacher is imperfect (\(\lambda < 1\)). Proofs for the theoretical analysis are provided in the \texttt{supplementary}. 
\begin{table*}[h]
\small
\centering
\renewcommand{\arraystretch}{1.2}
\vspace{-4pt}
\caption{Performance comparison on \textit{Cityscapes}$ \rightarrow$ \textit{Foggy Cityscapes} (C$\rightarrow$F),  
         \textit{Sim10k}$ \rightarrow$ \textit{Cityscapes} (S$\rightarrow$C), and  
         \textit{Kitti}$ \rightarrow$ \textit{Cityscapes} (K$\rightarrow$C). The best in each category is bold.}
\label{tab:c2fs2ck2c}
\resizebox{\linewidth}{!}{%
\begin{tabular}{@{}llccccccccc|cc@{}}
\toprule
 & & \multicolumn{9}{c|}{\textbf{C$\rightarrow$F}} & \textbf{S$\rightarrow$C} & \textbf{K$\rightarrow$C} \\
\cmidrule(lr){3-13}
\textbf{Category} & \textbf{Method} &
prsn & rider & car & truck & bus & train & mcycle & bicycle & \textbf{mAP} &
\textbf{AP Car} & \textbf{AP Car} \\
\midrule
\multirow[c]{1}{*}{\textbf{S}}
  & Source Only   
  & 29.3 & 34.1 & 35.8 & 15.4 & 26.0 & 9.09 & 22.4 & 29.7 & 25.2 & 32.0 & 33.9 \\
\midrule
\multirow[c]{4}{*}{\textbf{UDAOD}}
  & SSAL \citep{munir2021ssal}  \scriptsize \textcolor{gray!60}{(NeurIPS'21)} 
  & 45.1 & 47.4 & 59.4 & 24.5 & 50 & 25.7 & 26 & 38.7 & 39.6 & 51.8 & 45.6 \\ 
  & PT \citep{pt}  \scriptsize \textcolor{gray!60}{(ICML'22)} 
  & 40.2 & 48.8 & 59.7 & 30.7 & 51.8 & 30.6 & 35.4 & 44.5 & 42.7 & 55.1 & 60.2 \\
  & MTM \citep{weng2024mean}  \scriptsize \textcolor{gray!60}{(AAAI'24)} 
  & 51 & 53.4 & 67.2 & 37.2 & 54.4 & 41.6 & 38.4 & 47.7 & 48.9 & 58.1 & - \\
  & SEEN-DA \citep{li2025seen}  \scriptsize \textcolor{gray!60}{(CVPR'25)} 
  & 58.5 & 64.5 & 71.7 & 42 & 61.2 & 54.8 & 47.1 & 59.9 & 57.5 & 66.8 & 67.1 \\
\midrule
\multirow[c]{7}{*}{\textbf{SFOD}}
  & SFOD \citep{sfod} \scriptsize \textcolor{gray!60}{(AAAI’21)} 
  & 21.7 & 44.0 & 40.4 & 32.2 & 11.8 & 25.3 & 34.5 & 34.3 & 30.6 & 42.3 & 43.6 \\
  & SFOD-Mosaic \citep{sfod_mosaic}   \scriptsize \textcolor{gray!60}{(AAAI’21)} 
  & 25.5 & 44.5 & 40.7 & 33.2 & 22.2 & 28.4 & 34.1 & 39.0 & 33.5 & 42.9 & 44.6 \\
  & LODS \citep{lods}  \scriptsize \textcolor{gray!60}{(CVPR’22)} 
  & 34.0 & 45.7 & 48.8 & 27.3 & 39.7 & 19.6 & 33.2 & 37.8 & 35.8 & - & 43.9 \\
  & IRG \citep{irg}  \scriptsize \textcolor{gray!60}{(CVPR’23)} 
  & 37.4 & 45.2 & 51.9 & 24.4 & 39.6 & 25.2 & 31.5 & 41.6 & 37.1 & 45.2 & 46.9 \\
  & PETS \citep{pets}   \scriptsize \textcolor{gray!60}{(ICCV’23)} 
  & 42.0 & 48.7 & 56.3 & 19.3 & 39.3 & 5.5 & 34.2 & 41.6 & 35.9 & 57.8 & 47.0 \\
  & Simple-SFOD \citep{hao2024simplifying}  \scriptsize \textcolor{gray!60}{(ECCV’24)} 
  & 40.9 & 48 & 58.9 & 29.6 & 51.9 & 50.2 & 36.2 & 44.1 & 45.0 & 55.4 & 46.2 \\
  & SF-YOLO \citep{varailhon2024source}  \scriptsize \textcolor{gray!60}{(ECCV’24 w)} 
  & 47.4 & 50.6 & \textbf{64.1} & 26.0 & 49.7 & 32.5 & 25.9 & 44.1 & 42.5 & 57.7 & 49.4 \\
  & DRU \citep{khanh2024dynamic}  \scriptsize \textcolor{gray!60}{(ECCV’24)} 
  & \textbf{48.3} & \textbf{51.5} & 62.5 & 26.2 & 43.2 & 34.1 & 34.2 & \textbf{48.6} & 43.7 & 58.7 & 45.1 \\
  \cmidrule(lr){2-13}
  & \textbf{FALCON-SFOD (Ours)}   
  & 41.0 & 48.3 & 58.7 & \textbf{33.6} & \textbf{54.8} & \textbf{54.3} & \textbf{38.6} & 46.2 & \textbf{46.9} & \textbf{58.8} & \textbf{50.1} \\
\bottomrule
\end{tabular}}
\end{table*}

\section{Experiments}
\label{sec:experiments}

\textbf{Datasets and Metrics.}
We use five publicly available datasets covering four domain shift scenarios. Cityscapes~\citep{cordts2016cityscapes} is an urban street scene dataset comprising 5,000 finely annotated images collected from diverse cities and seasons, from which we use 2,925 images for training and 500 for validation. It includes eight categories: person, rider, car, truck, bus, train, motorcycle, and bicycle. Foggy Cityscapes~\citep{sakaridis2018semantic} extends Cityscapes by overlaying synthetic fog at three intensity levels (0.005, 0.01, and 0.02) to simulate poor visibility conditions. KITTI~\citep{geiger2013vision} is an autonomous driving dataset consisting of 7,481 real-world street scene training images. Sim10k~\citep{johnson2016driving} provides 10,000 synthetic urban scene images of cars, rendered from the video game Grand Theft Auto. BDD100k~\citep{yu2018bdd100k} is a large-scale dataset comprising 100,000 driving scene images captured across various weather conditions. Following existing works, we report mean average precision (mAP) at 0.5 IoU threshold. To demonstrate the efficacy of our method on extreme domain shifts, we use following datasets: PascalVOC~\citep{everingham2010pascal}, COCO~\citep{lin2014microsoft}, FLIR~\citep{flir2019adas}, and Clipart~\citep{inoue2018cross}.

\vspace{4pt}
\noindent \textbf{Implementation Details.}
\label{sec:implementation_details}
Our student model is trained using SGD with learning rates of 0.04 for source-only training and 0.0025 for target-domain adaptation, and a batch size of 4. The teacher model is updated via EMA with decay factor $\delta$ = 0.9996, using a confidence threshold 0.8. We use ~\cite{yuan2024open} for getting the class-agnostic binary foreground masks for SPAR loss. The training parameters are identical for every dataset. All experiments are run on one NVIDIA RTX A6000 GPU. Results in the paper use \cite{hao2024simplifying} as the base model unless specified otherwise. Similar to ~\cite{irg, hao2024simplifying}, we use 0.02 fog-level images from the Foggy Cityscapes for adaptation.

\subsection{Comparisons with State-of-the-art}
\textbf{Adaptation to Adverse Weather.} 
To evaluate the adverse weather domain shifts, we perform adaptation from Cityscapes to Foggy Cityscapes. Table \ref{tab:c2fs2ck2c} shows that our method achieves a state-of-the-art (sota) mAP of \textbf{46.9}\%, outperforming DRU~\cite{khanh2024dynamic} by $3.2\%$, SF-YOLO~\cite{varailhon2024source} by $4.9\%$, and Simple-SFOD~\cite{hao2024simplifying} by $1.9\%$. Especially, our method achieves notable improvements of 4.1\%, 2.9\%, and 2.4\% in the under-represented and challenging classes train, bus, and motorcycle, respectively, compared to the previous sota in those categories.

\noindent \textbf{Synthetic to Real-world.} 
We use the Sim10k dataset as the source-domain and the car category from Cityscapes as the target-domain (Table~\ref{tab:c2fs2ck2c} S$\rightarrow$C). Our method achieves a performance of \textbf{58.8}\%, beating Simple-SFOD by 3.4\% and SF-YOLO by 1.1\%. 

\noindent \textbf{Cross-camera Adaptation.} 
Table~\ref{tab:c2fs2ck2c} shows that in cross-camera adaptation scenarios (K$\rightarrow$C), our proposed approach achieves a performance of \textbf{50.1}\%, outperforming PETS~\cite{pets} by 3.1\% and DRU ~\cite{khanh2024dynamic} by 5\%. 

\noindent \textbf{Small-scale to Large-scale.}
We select Cityscapes as the source-domain and BDD100k as the target-domain to study this shift. Following \cite{pets, hao2024simplifying}, we focus on the seven categories shared with Cityscapes. As shown in Table \ref{tab:city2bdd}, our method achieves an mAP of \textbf{36.9}\%, beating Simple-SFOD~\cite{hao2024simplifying} by 2.6\%.

\noindent \textbf{Results on Extreme Shifts.}
To stress-test our method under severe domain shifts, we evaluate on three challenging transfers: i) Realistic to artistic data, ii) RGB to Thermal and iii) Thermal to RGB.
Tables \ref{tab:voc2clipart} and \ref{tab:flir_combined_results} show that our method consistently improves performance (by $\sim$2 mAP) under extreme domain shifts, underscoring its robustness.

\begin{table*}[h]
\small
\centering
\caption{Performance comparison on \textit{Cityscapes}$ \rightarrow$ \textit{BDD100k}. The best in each category is bold.}
\label{tab:city2bdd}
\renewcommand{\arraystretch}{1.2} 
\resizebox{0.85\linewidth}{!}{%
\begin{tabular}{@{}llcccccccc@{}}
\toprule
\textbf{Category} & \textbf{Method} &
truck & car & rider & person & motor & bicycle & bus & \textbf{mAP} \\
\midrule
\multirow{1}{*}{\textbf{S}}
  & Source Only
  & 9.9 & 51.5 & 17.8 & 28.7 & 7.5 & 10.8 & 7.6 & 19.1 \\
\midrule
\multirow{3}{*}{\textbf{UDAOD}}
  & SWDA \citep{swda}  \scriptsize \textcolor{gray!60}{(CVPR’19)}
  & 15.2 & 45.7 & 29.5 & 30.2 & 17.1 & 21.2 & 18.4 & 25.3 \\
  & CR-DA-Det \citep{cat_reg}  \scriptsize \textcolor{gray!60}{(CVPR’20)}
  & 19.5 & 46.3 & 31.3 & 31.4 & 17.3 & 23.8 & 18.9 & 26.9 \\
  & MTM \citep{weng2024mean}  \scriptsize \textcolor{gray!60}{(AAAI’24)}
  & 53.7 & 35.1 & 68.8 & 23.0 & 28.8 & 23.8 & 28.0 & 37.3 \\
\midrule
\multirow{8}{*}{\textbf{SFOD}}
  & SFOD \citep{sfod}   \scriptsize \textcolor{gray!60}{(AAAI’21)}
  & 20.4 & 48.8 & 32.4 & 31.0 & 15.0 & 24.3 & 21.3 & 27.6 \\
  & SFOD-Mosaic \citep{sfod_mosaic}   \scriptsize \textcolor{gray!60}{(AAAI’21)}
  & 20.6 & 50.4 & 32.6 & 32.4 & 18.9 & 25.0 & 23.4 & 29.0 \\
  & A$^2$SFOD \citep{a2sfod}   \scriptsize \textcolor{gray!60}{(AAAI’23)}
  & 26.6 & 50.2 & 36.3 & 33.2 & 22.5 & 28.2 & 24.4 & 31.6 \\
  & IRG \citep{irg}   \scriptsize \textcolor{gray!60}{(CVPR’23)}
  & 31.4  & 59.7 & 32.8 & 39.9 & 16.7 & 26.9 & 21.5 & 32.7 \\
  & PETS \citep{pets}  \scriptsize \textcolor{gray!60}{(ICCV’23)}
  & 19.3 & 62.4 & 34.5 & 42.6 & 17.0 & 26.3 & 16.9 & 31.3 \\
  & Simple-SFOD  \citep{hao2024simplifying}   \scriptsize \textcolor{gray!60}{(ECCV’24)}
  & 32 & 60 & 33.4 & 40.2 & 19.7 & 29.9 & 24.9 & 34.3 \\
  & DRU \citep{khanh2024dynamic}   \scriptsize \textcolor{gray!60}{(ECCV’24)}
  & 27.1 & \textbf{62.7} & \textbf{36.9} & \textbf{45.8} & 22.7 & 32.5 & 28.1 & 36.6 \\
  \cmidrule(lr){2-10}
  & \textbf{FALCON-SFOD (Ours)}
  & \textbf{32.6} & 59.8 & 34.0 & 40.0 & \textbf{25.7} & \textbf{35.7} & \textbf{30.5} & \textbf{36.9} \\
\bottomrule
\end{tabular}}
\end{table*}

\begin{table*}[h]
\centering
\renewcommand{\arraystretch}{1.2}
\caption{Performance comparision on Pascal VOC $\rightarrow$ Clipart.}
\label{tab:voc2clipart}
\resizebox{\linewidth}{!}{%
\setlength{\tabcolsep}{4pt}
\begin{tabular}{@{}l*{21}{c}@{}}
\toprule
\textbf{Method} &
Aero & Bicycle & Bird & Boat & Bottle & Bus & Car & Cat &
Chair & Cow & Table & Dog & Horse & Bike & Person & Plant &
Sheep & Sofa & Train & TV & \textbf{mAP} \\
\midrule
Baseline~\cite{hao2024simplifying}
& 24.1 & 59.3 & 28.9 & 22.3 & 29.9 & 59.4 & 41.2 & 11.4 & 37.8 & 15.7
& 28.3 & 13.1 & 39.3 & 57.1 & 53.4 & 41.9 & 15.7 & 21.5 & 34.5 & 36.8
& 33.6 \\
FALCON-SFOD (Ours)
& 26.5 & 60.6 & 31.1 & 24.9 & 30.8 & 62.4 & 41.6 & 14.1 & 40.0 & 17.1
& 29.4 & 15.5 & 39.9 & 58.4 & 55.6 & 43.2 & 16.1 & 23.6 & 38.7 & 37.5
& \textbf{35.5} \\
\bottomrule
\end{tabular}
}
\end{table*}

\begin{table}[h]
\centering
\renewcommand{\arraystretch}{1.2}
\caption{FLIR cross-domain detection results.}
\label{tab:flir_combined_results}
\resizebox{\linewidth}{!}{%
\begin{tabular}{@{}lcccccccc@{}}
\toprule
\multirow{2}{*}{\textbf{Method}} &
\multicolumn{4}{c}{\textbf{FLIR Visible $\rightarrow$ Infrared}} &
\multicolumn{4}{c}{\textbf{FLIR Infrared $\rightarrow$ COCO}} \\
\cmidrule(lr){2-5} \cmidrule(lr){6-9}
& person & bicycle & car & \textbf{mAP} & person & bicycle & car & \textbf{mAP} \\
\midrule
Basline~\cite{hao2024simplifying} & 59.8 & 42.3 & 68.1 & 56.7 & 25.2 & 12.7 & 20.3 & 19.4 \\
FALCON-SFOD (Ours) & 61.9 & 43.7 & 69.9 & \textbf{58.5} & 27.1 & 13.8 & 21.9 & \textbf{20.9} \\
\bottomrule
\end{tabular}}
\end{table}

\begin{table*}[h]
    \centering
    \caption{\textbf{Ablation}: Study of different components of our model on two domain shift scenarios.}
    \vspace{-4pt}
    \label{tab:ablations}
    \scalebox{0.8}{
    \begin{tabular}{c|c|c|c|c|c|c|c|c|c|c|c|c}
        \toprule
        \multirow{2}{*}{\textbf{Method}}
        & \multicolumn{2}{c|}{\textbf{Components}}
        & \multicolumn{9}{c|}{\textbf{C $\rightarrow$ F}}
        & \multicolumn{1}{c}{\textbf{S $\rightarrow$ C}} \\
        \cmidrule(lr){2-3}
        \cmidrule(lr){4-12}
        \cmidrule(lr){13-13}
        & \textbf{SPAR} & \textbf{IRPL}
        & prsn & rider & car & truck & bus & train & mcycle & bicycle & \textbf{mAP}
        & \textbf{AP Car} \\
        \midrule
        Baseline & \xmark & \xmark & 40.9 & 48.0 & 58.9 & 29.6 & 51.9 & 50.2 & 36.2 & 44.1 & 45.0 & 55.4 \\
        + SPAR   & \cmark & \xmark & 41.0 & 48.2 & 58.9 & 31.9 & 52.7 & 53 & 37.4 & 45.8 & 46.1 & 57.5 \\
        + IRPL   & \xmark & \cmark & 40.9 & 48.1 & 58.6 & 31.3 & 52.4 & 52.6 & 36.9 & 45.4 & 45.8 & 56.8 \\
        FALCON-SFOD (Ours)     & \cmark & \cmark & 41.0 & 48.3 & 58.7 & 33.6 & 54.8 & 54.3 & 38.6 & 46.2 & \textbf{46.9} & \textbf{58.8} \\
        \bottomrule
    \end{tabular}}
    \vspace{-12pt}
\end{table*}

\subsection{Ablation Studies}
\label{sec:ablations}
\noindent \textbf{Ablation on different components of our approach.}
Table \ref{tab:ablations} demonstrates the impact of the proposed components SPAR and IRPL across two domain shift scenarios. Adding SPAR alone achieves a gain of +1.1\% and +2.1\% on the two domain shifts, respectively. IRPL achieves a $\sim$+1.0\% increase in the performance. Combining these components yields the best results, enhancing mAP by +1.9\% and AP by +3.4\% in the respective domain shifts. These results confirm the complementary nature of SPAR and IRPL in enhancing the detection performance. 

\noindent \textbf{Ablation on different binary masks in SPAR.}
Table \ref{tab:binarymask} presents an ablation study evaluating different types of binary masks used in SPAR. Using source maps obtained from the source-trained model provides a clear improvement, showing the benefit of incorporating spatial priors within SPAR. Further enhancements are observed when integrating binary masks from recent methods such as GSAM, ESC-Net, and OVSAM, with OVSAM achieving the highest accuracy on all transfer tasks (C→F, S→C, K→C). These results demonstrate that SPAR consistently benefits from richer and more semantically aligned binary masks.

\begin{table}[h]
\small
\centering
\renewcommand{\arraystretch}{1.2}
\caption{\textbf{Ablation}: Study on the choice of binary masks in SPAR. Source maps refer to the thresholded channel-mean maps obtained from the source-trained model.}
\vspace{-4pt}
\label{tab:binarymask}
\resizebox{\linewidth}{!}{%
\begin{tabular}{@{}llccc@{}}
\toprule
\textbf{Method} & \textbf{Binary-Mask in SPAR} & \textbf{(C$\rightarrow$F)} & \textbf{(S$\rightarrow$C)} & \textbf{(K$\rightarrow$C)} \\
\midrule
Baseline & -- & 45.0 & 55.4 & 46.2 \\
\midrule
\multirow{4}{*}{Ours} 
& Source maps & 45.8 & 56.5 & 47.5 \\
& GSAM~\citep{ren2024grounded} \scriptsize \textcolor{gray!60}{(arXiv’24)} & 46.2 & 57.2 & 48.4 \\
& ESC-Net~\citep{lee2025effective} \scriptsize \textcolor{gray!60}{(CVPR’25)} & 46.5 & 57.8 & 49.3 \\
& OVSAM~\citep{yuan2024open} \scriptsize \textcolor{gray!60}{(ECCV’24)} & \textbf{46.9} & \textbf{58.8} & \textbf{50.1} \\
\bottomrule
\end{tabular}}
\vspace{-12pt}
\end{table}

\begin{figure}[h]
    \centering
    \vspace{-6pt}
\includegraphics[width=\linewidth]{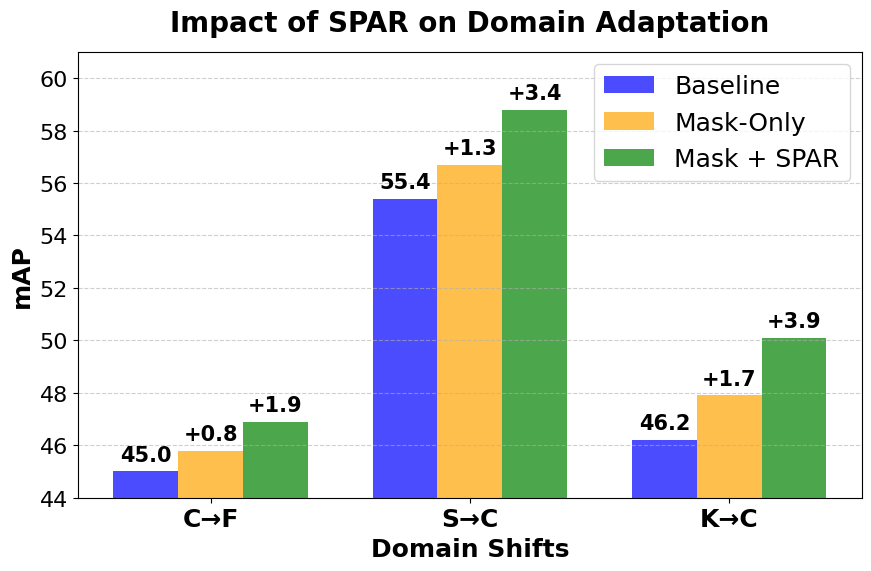}
    \caption{Study of the impact of SPAR on source-free adaptation.}
    \label{fig:spar_impact}
\vspace{-6pt}
\end{figure}

\noindent \textbf{Impact of SPAR Loss.}
Figure \ref{fig:spar_impact} illustrates performance across three domain shifts under different configurations: Baseline, Mask-Only, and Mask + SPAR.
In the Mask-Only setup, we employ OVSAM-derived binary masks solely for pseudo-label filtering, excluding the SPAR loss. Alongside score-based filtering, we compute the IoU between the teacher model’s predictions and the foreground regions in the binary masks, applying an IoU threshold of 0.5 for selection. As shown, this yields only moderate improvements compared to the Mask + SPAR configuration, which integrates our proposed SPAR loss with the OV-SAM masks. The observed performance gap demonstrates that the gains are not merely due to improved pseudo-label selection using external masks. Instead, the SPAR loss itself enhances feature-foreground alignment, fostering more meaningful and generalizable representations within the source-free adaptation framework.

\noindent \textbf{Ablation on different components of IRPL.}
Table \ref{tab:ablations_irpl} presents an ablation study highlighting the impact of different components in IRPL (Equation \ref{eq:irpl}). The peak-adjust transform ($p'_{k}$)~\cite{wang2024epsilon} provides moderate improvement when directly applied to object detection, yielding 55.9 AP. However, better performance is achieved when it is complemented with: i) foreground-background weighting ($\mathbf{w_{\hat{c}}}$), which addresses the inherent imbalance between object and background regions and ii) entropy regularization ($\mathbf{D_{\mathrm{KL}}}$), which mitigates head-class dominance. Together, these components lead to the highest AP of 56.8, demonstrating the effectiveness of IRPL in SFOD. 

\begin{table}[h]
    \centering
    \caption{\textbf{Ablation}: Study of different components of IRPL.}
    \vspace{-2pt}
    \label{tab:ablations_irpl}
    \scalebox{0.95}{
    \begin{tabular}{c|c|c|c|c}
        \toprule
        \multirow{2}{*}{\textbf{Method}}
        & \multicolumn{3}{c|}{\textbf{Components}}
        & \multicolumn{1}{c}{\textbf{S $\rightarrow$ C}} \\
        \cmidrule(lr){2-4}
        \cmidrule(lr){5-5}
        & $\mathbf{p'_{k}}$ & $\mathbf{w_{\hat{c}}}$ & $\mathbf{D_{\mathrm{KL}}}$
        & \textbf{AP Car} \\
        \midrule
        Baseline & - & - & - & 55.4 \\
        \midrule
        + peak-adjust  & \cmark & \xmark & \xmark & 55.9 \\
        + fg-bg weighting & \cmark & \cmark & \xmark & 56.3 \\
        + entropy (Ours)    & \cmark & \cmark & \cmark & \textbf{56.8} \\
        \bottomrule
    \end{tabular}}
    \vspace{-2pt}
\end{table}

\noindent \textbf{Analysis on long-tail class imbalance problem.}
As seen in Table~\ref{tab:class-imbalance}, our method achieves clear advantages on under-represented and challenging categories. The largest AP gains occur in tail classes with very few target instances, such as \emph{train} ($+4.1$), \emph{truck} ($+4.0$), and \emph{bus} ($+2.9$), while common classes like \emph{car} and \emph{person} show marginal changes ($\leq 0.2$ AP). Quantitatively, the Pearson correlation between $\log_{10}$(class frequency) and AP improvement ($\Delta$) is strongly negative ($r_\Delta = -0.90$), confirming that our method systematically benefits difficult categories the most. These results demonstrate that IRPL effectively mitigates long-tail imbalance by enhancing performance on difficult, low-frequency classes without degrading head-class accuracy.

\vspace{-6pt}
\begin{table}[h]
\small
\centering
\caption{Per-category AP$_{50}$ results on the target-domain. We report baseline~\cite{hao2024simplifying} performance, performance with our method, and the improvement ($\Delta$).}
\label{tab:class-imbalance}
\scalebox{0.98}{
\begin{tabular}{lcccc}
\toprule
\textbf{Category} & \textbf{Target instances} 
& \textbf{Baseline} & \textbf{Ours} & $\Delta$ \\
\midrule
person      & 3\,419 & 40.9 & 41.0 & +0.1 \\
rider       &   556  & 48.0 & 48.3 & +0.3 \\
car         & 4\,667 & 58.9 & 58.7 & –0.2 \\
truck       &    93  & 29.6 & 33.6 & +4.0 \\
bus         &    98  & 51.9 & 54.8 & +2.9 \\
train       &    23  & 50.2 & 54.3 & +4.1 \\
motorcycle  &   149  & 36.2 & 38.6 & +2.4 \\
bicycle     & 1\,175 & 44.1 & 46.2 & +2.1 \\
\midrule
mAP         &   ---  & 45.0 & \textbf{46.9} & +1.9 \\
\bottomrule
\end{tabular}}
\vspace{-8pt}
\end{table}

In the \texttt{supplementary material}, we provide: (i) pseudo-code of the proposed method, (ii) results on transformer-based SFOD methods, (iii) additional qualitative examples, (iv) compute analysis, and (v) hyperparameter studies, among other details.
\vspace{-6pt}
\section{Conclusion}
In this work, we observed that domain shift weakens the detector's object-focus in the feature space and proposed FALCON-SFOD, a theoretically grounded framework that effectively mitigates this challenge. Our method leverages foundation priors through the proposed SPAR loss to suppress spurious background activations and promote structured, foreground-focused features. Complementing this, IRPL enhances adaptation stability by addressing object imbalance and residual label noise. A comprehensive theoretical analysis provides tighter risk bounds that formally connect our objectives to improved localization and classification robustness. Extensive experiments across diverse benchmarks and domain shift scenarios demonstrate that FALCON-SFOD achieves competitive performance, consistently surpassing existing SFOD approaches.

{
    \small
    \bibliographystyle{ieeenat_fullname}
    \bibliography{main}
}

\appendix
\onecolumn
\pagenumbering{arabic}
\setcounter{page}{1}

\setcounter{section}{0}
\setcounter{table}{0}
\setcounter{figure}{0}
\setcounter{equation}{0}

\renewcommand{\thesection}{A.\arabic{section}}
\renewcommand{\thetable}{A.\arabic{table}}
\renewcommand{\thefigure}{A.\arabic{figure}}
\renewcommand{\theequation}{A.\arabic{equation}}

\newcommand{\ToCEntry}[3]{%
  \ifcase#1
    \noindent\ref{#3}. #2\hspace{1.5em}\dotfill\hspace{1.5em}\pageref{#3} \\ 
  \or
    \noindent\hspace*{2em}\ref{#3}. #2\hspace{1.5em}\dotfill\hspace{1.5em}\pageref{#3} \\ 
  \else
    \noindent\ref{#3}. #2\hspace{1.5em}\dotfill\hspace{1.5em}\pageref{#3} \\ 
  \fi
}

\begin{center} \section*{\Large Appendix for \\ Foundation Model Priors Enhance Object Focus in Feature Space for Source-Free Object Detection} \end{center}

\section*{Contents}
\ToCEntry{0}{Algorithm for the Proposed Method}{sec:pseudo-code}
\ToCEntry{0}{Architecture-generality of FALCON-SFOD}{sec:architecture_general}
\ToCEntry{0}{Reproducibility and Implementation Details}{sec:reproducibility}
\ToCEntry{1}{Hyperparameter Analysis}{sec:hyperparameter_analysis}
\ToCEntry{1}{Results across Random Seeds}{sec:random_seeds}
\ToCEntry{0}{Proofs for the Lemmas and Theorems}{sec:proofs}
\ToCEntry{0}{Run time and Memory Analysis}{sec:compute_analysis}
\ToCEntry{0}{Additional Qualitative Results}{sec:additional_qual}
\hrule

\section{Algorithm of FALCON-SFOD}
\label{sec:pseudo-code}

\begin{algorithm}[H]
\caption{Training Loop of the proposed method}
\label{alg:sfdaod}
\begin{algorithmic}[1]
\Require Teacher \(h^{\mathrm{te}}\), student \(h^{\mathrm{st}}\);
         target images \(\mathcal{X}_t\)
\Require 
optimiser \(\mathrm{Opt}(\cdot)\), hyperparameters \\
Pre-compute binary foreground masks \(A_{\mathrm G}\)  
\For{\textbf{each} mini-batch \(\mathcal{B}\subset\mathcal{X}^t\)}
    \State \textbf{Augment}: obtain weak/strong views
           \((\tilde{x}_i^{t},\bar{x}_i^{t})\) for every \(x_i^t\in\mathcal{B}\)
    \State \textbf{Teacher forward}:
           \( \tilde{\mathcal{Y}}_i^t
            \leftarrow h^{\mathrm{te}}(\tilde{x}_i^{t})\)
    \State \textbf{Pseudo-labels}:
           \( \hat{\mathcal{Y}}_i^t = \{(\hat{b}_{ij}, \hat{c}_{ij})\}
            \leftarrow\text{Filter}(\tilde{\mathcal{Y}}_i^t)\)
    \State \textbf{Student forward}:
           \((\mathbf p_{i},\mathbf b_{i})
            \leftarrow h^{\mathrm{st}}(\bar{x}_i^{t})\)
    
    \State Apply Eq.\,\ref{eq:eps_softmax} to
           \(\mathbf p_{i}\) to obtain
           \(\mathbf p'_i\)
    \State \(w_{\hat{c}}\leftarrow
           \begin{cases}
             w_{\mathrm{fg}}, & \hat{c} \text{ is foreground},\\
             w_{\mathrm{bg}}, & \text{otherwise}
           \end{cases}\)
    \State Compute \(\mathcal{L}_{\mathrm{IRPL}}\)
           via Eq.\,\ref{eq:irpl}
    \State Compute student mean maps
           \(A_{\mathrm S}(x_i^{t})=
           \text{mean-channel}\bigl(g^{\mathrm st}(x_i^{t})\bigr)\) 
    \State Compute \(\mathcal{L}_{\mathrm{SPAR}}\)
           via Eq.\,\ref{eq:spar}
    \State Compute standard detection localization loss
           \(\mathcal{L}_{reg}\)
    \State Aggregate and update: 
           \(\theta^{\mathrm S}\leftarrow
           \mathrm{Opt}\bigl(\theta^{\mathrm st},
           \nabla_{\theta^{\mathrm st}}
           (\mathcal{L}_{\mathrm{IRPL}}
            +\mathcal{L}_{\mathrm{SPAR}}
            +\mathcal{L}_{reg})\bigr)\)
\EndFor
\State \textbf{Update teacher}: $\theta^{te} \leftarrow
        \delta \theta^{te}
        +(1-\delta)\theta^{st}$ \\
\Return Adapted teacher \(h^{\mathrm{te}}\)
\end{algorithmic}
\end{algorithm}

\begin{table*}[h]
\small
\centering
\vspace{-4pt}
\caption{Performance comparison of our method when integrated into different detection architectures on \textit{Cityscapes}$ \rightarrow$ \textit{Foggy Cityscapes} (C$\rightarrow$F),  
         \textit{Sim10k}$ \rightarrow$ \textit{Cityscapes} (S$\rightarrow$C), and  
         \textit{Kitti}$ \rightarrow$ \textit{Cityscapes} (K$\rightarrow$C). Note that FALCON-SFOD consistently improves perforamance across architectures and domain shifts.}
\label{tab:agnostic_c2fs2ck2c}
\resizebox{\linewidth}{!}{%
\begin{tabular}{@{}lccccccccc|cc@{}}
\toprule
 & \multicolumn{9}{c|}{\textbf{C$\rightarrow$F}} & \textbf{S$\rightarrow$C} & \textbf{K$\rightarrow$C} \\
\cmidrule(lr){2-12}
\textbf{Method}  &
prsn & rider & car & truck & bus & train & mcycle & bicycle & \textbf{mAP} &
\textbf{AP Car} & \textbf{AP Car} \\
\midrule
IRG \citep{irg}  \scriptsize \textcolor{gray!60}{(CVPR’23)} 
& 37.4 & 45.2 & 51.9 & 24.4 & 39.6 & 25.2 & 31.5 & 41.6 & 37.1 & 45.2 & 46.9 \\
\textbf{+ (Ours)}   
& 37 & 45.9 & 51.7 & 30.2 & 44.7 & 30 & 32.9 & 40.6 & \textbf{39.0} & \textbf{49.1} & \textbf{49.8} \\
PETS \citep{pets}   \scriptsize \textcolor{gray!60}{(ICCV’23)} 
& 42.0 & 48.7 & 56.3 & 19.3 & 39.3 & 5.5 & 34.2 & 41.6 & 35.9 & 57.8 & 47.0 \\
\textbf{+ (Ours)}   
& 46.2 & 52.9 & 63.2 & 24 & 49.1 & 10.4 & 40.5 & 48.6 & \textbf{41.9} & \textbf{59.1} & \textbf{48.9} \\
DRU \citep{khanh2024dynamic}  \scriptsize \textcolor{gray!60}{(ECCV’24)} 
& 48.3 & 51.5 & 62.5 & 26.2 & 43.2 & 34.1 & 34.2 & 48.6 & 43.7 & 58.7 & 45.1 \\
\textbf{+ (Ours)}   
& 49.1 & 52.0 & 63.7 & 29.4 & 45.2 & 36.7 & 37.5 & 49.8 & \textbf{45.4} & \textbf{60.5} & \textbf{48.2} \\
\bottomrule
\end{tabular}}
\vspace{-6pt}
\end{table*}

\section{Architecture-generality of FALCON-SFOD}
\label{sec:architecture_general}
To assess the architectural generality of our framework, we integrate FALCON-SFOD with multiple representative detection methods, including both conventional Faster R-CNN-based pipelines~\cite{irg, pets} and transformer-based detectors such as Deformable DETR~\cite{khanh2024dynamic}. As shown in Table \ref{tab:agnostic_c2fs2ck2c}, incorporating our proposed SPAR and IRPL modules consistently improves performance across architectures and domain-shift settings. Notably, the observed gains are obtained without modifying the detector structure or adding inference-time cost, underscoring the plug-and-play nature of our design. These results demonstrate that FALCON-SFOD generalizes well beyond a single detector family, enhancing adaptation robustness across both convolutional and transformer architectures.

\section{Reproducibility and Implementation Details}
\label{sec:reproducibility}
As described in Section \ref{sec:architecture_general}, FALCON-SFOD can be seamlessly integrated into a variety of detector architectures, consistently yielding performance gains across settings. For a fair comparison, we preserve all training parameters such as batch size, number of epochs, learning rate, optimization schedule, and data preprocessing pipeline from the corresponding baseline models, applying FALCON-SFOD as an additional adaptation module without altering the underlying training configuration. For the newly introduced hyperparameters, Section \ref{sec:hyperparameter_analysis} demonstrates that a single configuration performs consistently across benchmarks, indicating that FALCON-SFOD exhibits strong robustness and low sensitivity to hyperparameter choices. 

\subsection{Hyperparameter Analysis}
\label{sec:hyperparameter_analysis}
\textbf{Spatial Prior-Aware Regularization.} SPAR enforces object-focused feature learning by aligning the student’s channel-mean activation map with class-agnostic foreground priors. It achieves this by combining mean $\ell_1$ and Dice terms with default weights $\lambda_1$ = 1 and $\lambda_2$ = 2 selected via a coarse sweep (see Table \ref{tab:spar_sweep_cf}) and kept fixed across all experiments. The $\ell_1$ loss enforces pixel-wise agreement, ensuring accurate correspondence in activation magnitudes, while the Dice term complements it by emphasizing overlap and boundary coherence, preventing degenerate solutions where the map matches in average value but misses the overall object shape. Empirically, this combination leads to more stable optimization and consistently higher performance across domain shifts compared to using either term alone. As shown in Table \ref{tab:spar_sweep_cf}, the mAP varies by less than 1 point across the entire $(\lambda_1,\lambda_2)\in(0,4)$ range, indicating that SPAR is largely insensitive to its weighting coefficients. We therefore fix a single setting $(\lambda_1{=}1,\lambda_2{=}2)$ for all experiments, confirming that the method does not depend on careful tuning for stable performance. 

\begin{table}[h]
  \centering
  \caption{SPAR $\lambda_1$/$\lambda_2$ sweep on C$\rightarrow$F. Best mAP in \textbf{bold}. \textbf{Note}: With mAP variation under one point, SPAR demonstrates strong stability with respect to its weighting coefficients.}
  \label{tab:spar_sweep_cf}
  \setlength{\tabcolsep}{6pt}
  \begin{tabular}{lcccccc}
    \toprule
    & \multicolumn{6}{c}{$\lambda_2$} \\
    \cmidrule(lr){2-7}
    $\lambda_1$ & 0 & 1 & 2 & 3 & 4 \\
    \midrule
    0 & 45 & 45.6 & 45.6 & 45.3 & 45.5 &  \\
    1 & 45.4 & 45.8 & \textbf{46.1} & 45.7 & 45.6 &  \\
    2 & 45.3 & 45.4 & 45.7 & 45.6 & 45.9 &  \\
    3 & 45.4 & 45.5 & 45.4 & 45.7 & 45.6 &  \\
    4 & 45.2 & 45.3 & 45.3 & 45.4 & 45.6 &  \\
    \bottomrule
  \end{tabular}
  \vspace{4pt}
\end{table}

\noindent \textbf{Imbalance-aware Noise-Robust Pseudo-Labeling.} IRPL addresses the dual challenges of label noise and foreground-background imbalance inherent in source-free object detection. Unlike the standard cross-entropy objective, which can be dominated by mislabeled high-confidence predictions, IRPL utilizes a \textit{peak-adjust transform} that moderates the student’s logits by adding a large margin $m$ to the most confident class and renormalizing, thereby dampening gradients for easy, clean samples while preserving full corrective signals for uncertain or mislabeled ones.  This mechanism provides intrinsic robustness to noisy pseudo-labels. To further stabilize learning under imbalance, IRPL combines this with \textit{foreground-background weighting} ($w_{\mathrm{fg}}, w_{\mathrm{bg}}$) and a mild \textit{entropy regularization} term weighted by $\gamma$, which discourages over-confident head-class predictions. 
Following the theoretical insights from \cite{wang2024epsilon}, we treat the coefficients $\alpha$ and $\beta$ in
\[
L = \alpha L_{\epsilon} + \beta L_{\text{symmetric}}
\]
as scaling factors rather than critical hyperparameters. Lemma 3 in \cite{wang2024epsilon} proves that the excess-risk bound of $\alpha L_{\epsilon} + \beta L_{\text{symmetric}}$ is identical to that of $\alpha L_{\epsilon}$ alone, implying that $\alpha$ and $\beta$ affect only the gradient magnitude rather than the theoretical robustness. Hence, we fix them following \cite{wang2024epsilon} instead of tuning.

\begin{figure}
    \centering
    \includegraphics[width=0.7\linewidth]{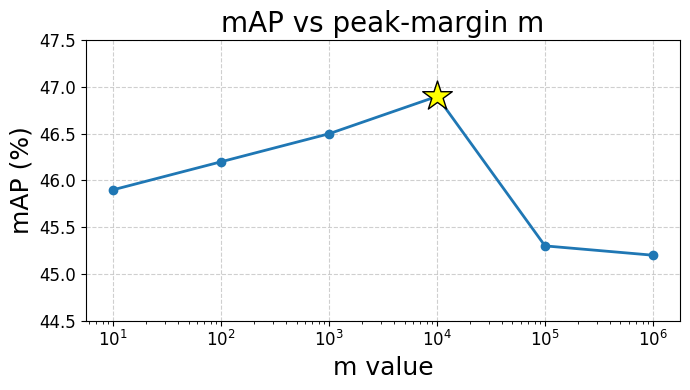}
    \caption{Performance on C $\rightarrow$ F with different m values used in the IRPL loss.}
    \label{fig:map_vs_m}
    \vspace{4pt}
\end{figure}

\vspace{4pt}
\begin{itemize}[itemsep=3pt]
    \item \textbf{$\alpha = 0.1$, $\beta = 2$.} These values balance the stronger gradients from the Cross-Entropy term and the weaker but noise-robust gradients from the symmetric component, keeping both contributions numerically comparable during training.
    \item \textbf{$m = 10^{4}$.} We swept $m \in \{10, 10^2, 10^3, 10^4, 10^5, 10^6\}$ (see Figure \ref{fig:map_vs_m}) and found $10^{4}$ to yield the best trade-off between clean-data fitting and robustness to noisy pseudo-labels, consistent with the analysis that larger $m$ produces more one-hot-like outputs but can lead to underfitting.
    \item $w_{fg} = 2$, $w_{bg} = 1$. Foreground regions are twice-weighted to emphasize object features over background context during imbalance-aware reweighting. However, we observe that excessively increasing $w_{fg}$ does not further improve performance, as it can overemphasize limited foreground samples and hinder domain alignment with background features.
    \item \textbf{$\gamma = 0.01$.} A mild entropy regularizer; larger values caused over-smoothing, while smaller values provided negligible regularization.
\end{itemize}

All values were chosen for stability and consistency across benchmarks rather than dataset-specific optimization. This configuration was empirically robust across all experiments without further tuning.

\subsection{Results across Random Seeds}
\label{sec:random_seeds}
\noindent\textbf{Statistical Summary.} As shown in Table~\ref{tab:c2f_seeds}, the mAP values for the three random seeds are 46.9, 46.7, and 47.4. 
The mean and standard deviation are computed as:
\[
\text{Mean} = 47.0, \quad \text{Std} = 0.29.
\]

\vspace{4pt}
\noindent\textbf{Discussion.} 
While existing SFOD works largely report single-run results, Table~\ref{tab:c2f_seeds} provides performance across three random seeds to evaluate robustness and stability. 
When integrated into Simple-SFOD~\cite{hao2024simplifying}, our FALCON-SFOD consistently improves performance on the challenging 
\textit{Cityscapes}$\rightarrow$\textit{Foggy Cityscapes} domain shift. 
Across seeds 42, 123, and 9999, our approach achieves an average mAP of \textbf{47.0~$\pm$~0.29}, 
demonstrating both a clear improvement over the Simple-SFOD baseline (45.0~mAP) and strong reproducibility. 
The small variance across runs indicates that FALCON-SFOD’s gains are stable and not sensitive to random initialization.

\begin{table*}[h]
\scriptsize 
\centering
\vspace{-6pt}
\caption{Performance comparison of our method (three random seeds) when integrated into Simple-SFOD~\cite{hao2024simplifying} on \textit{Cityscapes}$ \rightarrow$ \textit{Foggy Cityscapes} (C$\rightarrow$F). 
FALCON-SFOD shows consistent improvements and robustness across different runs.}
\label{tab:c2f_seeds}
\resizebox{0.95\linewidth}{!}{
\begin{tabular}{@{}lccccccccc@{}}
\toprule
 & \multicolumn{9}{c}{\textbf{C$\rightarrow$F}} \\
\cmidrule(lr){2-10}
\textbf{Method}  &
prsn & rider & car & truck & bus & train & mcycle & bicycle & \textbf{mAP} \\
\midrule
Simple-SFOD~\citep{hao2024simplifying}  \scriptsize \textcolor{gray!60}{(ECCV’24)} 
  & 40.9 & 48.0 & 58.9 & 29.6 & 51.9 & 50.2 & 36.2 & 44.1 & 45.0 \\
\textbf{+ (Ours, Seed 42)} 
  & 41.0 & 48.3 & 58.7 & 33.6 & 54.8 & 54.3 & 38.6 & 46.2 & 46.9 \\
\textbf{+ (Ours, Seed 123)} 
  & 41.3 & 48.1 & 59.2 & 32.7 & 54.2 & 53.5 & 38.4 & 46.3 & 46.7 \\
\textbf{+ (Ours, Seed 9999)} 
  & 41.6 & 48.6 & 59.5 & 34.1 & 55.1 & 54.9 & 39.1 & 46.8 & 47.4 \\
\bottomrule
\end{tabular}}
\vspace{-8pt}
\end{table*}

\section{Proofs for the Lemmas and Theorems}
\label{sec:proofs}

\textbf{Proof for Lemma \ref{lem:ce-general}.}
\begin{proof}
For any $x$ and any true class $j\in\{0,\dots,K\}$ define the non-negative
loss vector
$\boldsymbol{\ell}(x)=\bigl(\ell_0(x),\ldots,\ell_K(x)\bigr)^{\!\top}$ with
$\ell_i(x)=-\log p_i(x)\ge0$.
Since $T_{jj}\ge\lambda$ we have
\begin{align}
  \ell_j(x)
  &= \frac{T_{jj}}{T_{jj}}\,\ell_j(x)
  \le \frac{1}{\lambda}\,T_{jj}\,\ell_j(x) \notag\\
  &\le \frac1{\lambda}\sum_{i=0}^{K}T_{ji}\,\ell_i(x).
\end{align}
Taking expectation under the joint $(x,c)$ and using
$\Pr[\hat c=i\mid c=j,x]=T_{ji}$ yields
\begin{align}
  R^{cls}_{\mathcal{D}^T,\;clean}(f^{st}_c)
  &= \mathbb{E}_{(x,c)}\bigl[\ell_c(x)\bigr] \notag\\
  &\le \frac1{\lambda}\;
  \mathbb{E}_{(x,c)}\!
       \Bigl[\sum_{i}T_{c i}\,\ell_i(x)\Bigr] \notag\\
  &= \frac1{\lambda}\;
  R^{cls}_{\mathcal{D}^T,\;noisy}(f^{st}_c).
\end{align}
This completes the proof of Lemma~\ref{lem:ce-general}.
\end{proof}

\noindent \textbf{Proof for Lemma \ref{lem:reg-missing}.}
\begin{proof}
We write the clean risk as the expectation over the two disjoint events
$M=1$ and $M=0$:
\begin{align}
  \|f^{st}_r(x)-b\|_1
  &= M\,\|f^{st}_r(x)-b\|_1 \notag\\
  &\quad+\;(1-M)\,\|f^{st}_r(x)-b\|_1.
\end{align}

\noindent\textit{Case $M=1$ (teacher matched the box):}
When $M=1$ there exists the pseudo-box $\hat b$ and by the triangle
inequality
\begin{align}
  \|f^{st}_r(x)-b\|_1
  &\le
  \|f^{st}_r(x)-\hat b\|_1
  +\|\hat b-b\|_1.
\end{align}

\noindent\textit{Case $M=0$ (teacher missed the box):}
With normalised coordinates $\|f^{st}_r(x)-b\|_1\le 2$ for all $x,b$,
hence
\begin{align}
  (1-M)\,\|f^{st}_r(x)-b\|_1
  \le 2\,(1-M).
\end{align}

Taking expectations over $\mathcal{D}^T$ and summing the two cases gives
\begin{align}
  R^{reg}_{\mathcal{D}^T,\;clean}(f^{st}_r)
  &\le
  \mathbb{E}\!\bigl[
      M\,\|f^{st}_r(x)-\hat b\|_1
  \bigr] \notag\\
  &\quad+\mathbb{E}\!\bigl[
      M\,\|\hat b-b\|_1
  \bigr]
  + 2\,\mathbb{E}[1-M].
\end{align}
Using the definitions
$R^{reg}_{\mathcal{D}^T,\;noisy}=\mathbb{E}[M\,\|f^{st}_r(x)-\hat b\|_1]$,
$\eta_{reg}=\mathbb{E}[M\,\|\hat b-b\|_1]$, and
$\zeta=\mathbb{E}[1-M]$, we recover~\eqref{eq:reg-bound-missing}.
\end{proof}

\noindent \textbf{Proof for Theorem \ref{thm:det-bound-general}.}
\begin{proof}
Starting from the decomposition
\begin{align}
  R^{det}_{\mathcal{D}^T}
  &= R^{cls}_{{\mathcal{D}^T},clean}
   +R^{reg}_{{\mathcal{D}^T},clean},
\end{align}
(cf.\ Eq.~\ref{det_risk}),
we apply Lemma~\ref{lem:ce-general} to the classification term and
Lemma~\ref{lem:reg-missing} to the regression term:
\begin{align}
  R^{cls}_{{\mathcal{D}^T},clean}(f_c^{st})
  &\le \tfrac1{\lambda}\,R^{cls}_{{\mathcal{D}^T},noisy}(f_c^{st}),\\
  R^{reg}_{{\mathcal{D}^T},clean}(f_r^{st})
  &\le R^{reg}_{{\mathcal{D}^T},noisy}(f_r^{st})
  +\eta_{reg} + 2\,\zeta.
\end{align}
Adding the two inequalities gives~\eqref{eq:det-bound-general}.
\end{proof}

\noindent \textbf{Proof for Theorem \ref{thm:epsilon-excess}.}
\begin{proof}
Inspired from \citep{wang2024epsilon}, we can write
\begin{align}
  R^{\text{cls}}_{\mathcal D^{T},\text{clean}}\!\bigl(f_\eta^*\bigr)
  \le 2\delta+\frac{2w\,\delta}{a}.
\end{align}
Combining this with the regression bound that accommodates missed boxes
(Lemma~\ref{lem:reg-missing}),
\begin{align}
  R^{\text{reg}}_{\mathcal D^{T},\text{clean}}(f^{st}_r)
  \le
  R^{\text{reg}}_{\mathcal D^{T},\text{noisy}}(f^{st}_r)
  +\eta_{\text{reg}}+2\zeta,
\end{align}
and using the decomposition
$R^{\det}=R^{\text{cls}}_{\text{clean}}+R^{\text{reg}}_{\text{clean}}$
yields inequality~\eqref{eq:epsilon-excess-det}.
\end{proof}

\section{Run time and Memory Analysis}
\label{sec:compute_analysis}
Tables~\ref{tab:gsam_time} and~\ref{tab:gsam_mem} report the cost of integrating GSAM~\citep{ren2024grounded} into our pipeline. 
The time overhead is marginal: the offline GSAM pass completes in $1\,050$\,s ($\sim$17\,min), which is only $3.8\%$ of the $28\,000$\,s required for our model training (and only $5.2\%$ relative to the baseline). 
When added to the full pipeline, the end-to-end wall-clock increases only slightly ($28\,084$\,s $\rightarrow$ $29\,134$\,s), well within the typical run-to-run variability of large-scale training. 
The memory overhead is also short-lived: GSAM peaks at $18$\,GB only during its 17-minute preprocessing, while training itself never exceeds $9.6$\,GB. 
Since these stages do not overlap, the entire procedure fits comfortably on a single $24$-$48$\,GB GPU without any modification to training. 
Moreover, the amortization cost is small: GSAM masks are generated once per target split and can be cached for reuse in all subsequent experiments. 
The extraction step is fully parallel across images, so on a multi-GPU node the elapsed time approaches standard data-loading latency. 
\textit{For completeness, we note that ESC-Net~\citep{lee2025effective} and OV-SAM~\citep{yuan2024open} are both lighter than Grounded-SAM in parameter size, and therefore require less compute and memory}; we report GSAM values here since it represents the most demanding case among the three. 
Overall, even with GSAM enabled, the complete adaptation run finishes in under $8$ hours and $<18$\,GB peak memory on a single RTX A6000, confirming that the footprint remains modest.

\begin{table*}[h]
\centering
\renewcommand{\arraystretch}{1.2}
\caption{Run time comparison between baseline and our method on C $\rightarrow$ F domain shift. Note that the time taken for offline generation of foreground-masks is negligible compared to the training time. Also, the inference time remains the same as the baseline.}
\label{tab:gsam_time}
\resizebox{\linewidth}{!}{%
\begin{tabular}{@{}lcccc@{}}
\toprule
\textbf{Setting} & \textbf{Offline GSAM time (1000s)} & 
\textbf{Training time (1000s)} & \textbf{Test time (s)} & \textbf{End-to-end time (1000s)} \\
\midrule
IRG \cite{irg} & -- & 20 & 84 & 20.08 \\
IRG + Ours & -- & 28 & 84 & 28.08 \\
IRG + Ours + GSAM pre-processing cost & 1.050 & 28 & 84 & 29.13 \\
\bottomrule
\end{tabular}}
\end{table*}

\begin{table*}[h]
\centering
\renewcommand{\arraystretch}{1.2}
\caption{Memory usage comparison between baseline and our method on C $\rightarrow$ F domain shift. Note that the offline foreground-mask generation is short-lived and both the train-time and test-time memory remains same as the baseline.}
\label{tab:gsam_mem}
\resizebox{\linewidth}{!}{%
\begin{tabular}{@{}lccc@{}}
\toprule
\textbf{Setting} & \textbf{Offline GSAM peak mem (GB)} &
\textbf{Training peak mem (GB)} & \textbf{Stage-wise peak mem (GB)} \\
\midrule
IRG \cite{irg} & -- & 6.9 & 6.9 \\
IRG + Ours & -- & 9.6 & 9.6 \\
IRG + Ours + GSAM pre-processing cost & 18.4 & 9.6 & 18.4 \\
\bottomrule
\end{tabular}}
\end{table*}

\begin{figure*}[h]
    \centering
    \includegraphics[width=\linewidth, height=8cm]{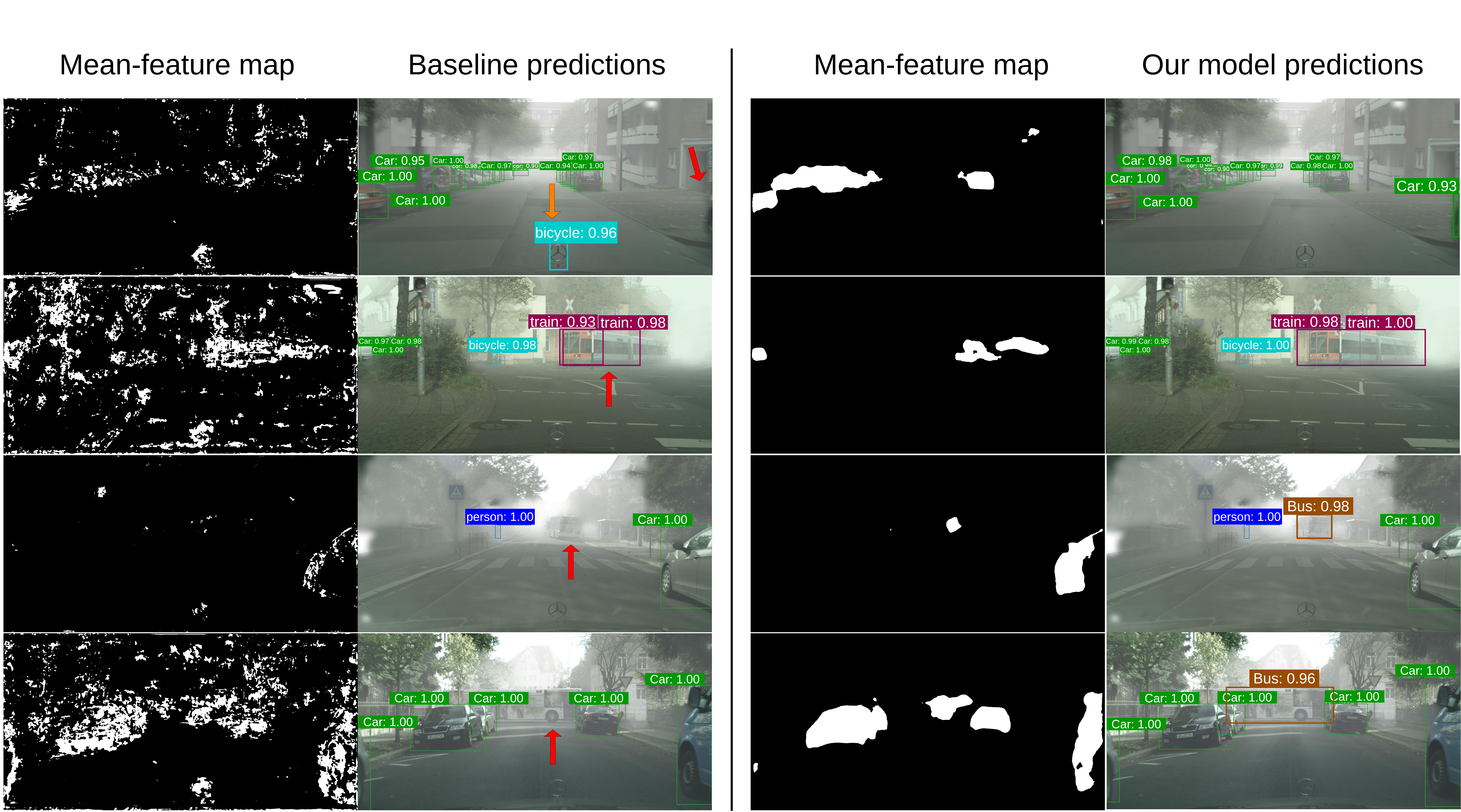}
    \vspace{-8pt}
    \caption{\textbf{Additional qualitative results.} Four examples from the Foggy Cityscapes~\citep{sakaridis2018semantic} target set. Mean-feature map is obtained from taking the channel-mean from the last layer of the student's backbone and thresholding at 0.6.\textbf{Left}: Baseline model ~\citep{hao2024simplifying} produces spurious background activations, leading to missed detections or localization errors (\textcolor{red}{red arrows}) and false positives (\textcolor{orange}{orange arrows}). \textbf{Right}: Our method effectively suppresses both feature-space confusion and class-label noise, resulting in clear activations and more accurate classification and object localization. \emph{Zoom in for best view}.}
\label{fig:additional_qualitative}
\end{figure*}

\section{Additional Qualitative Results}
\label{sec:additional_qual}
Figure \ref{fig:additional_qualitative} presents additional qualitative examples comparing the baseline method \citep{hao2024simplifying} and our proposed method on the Foggy Cityscapes dataset. Each row corresponds to one scene, with the first two columns illustrating the baseline's thresholded mean-channel feature maps and predicted detections, and the last two columns showing the same representations from our method. The mean-channel map is obtained by taking the mean along the channel dimension of the last layer of the backbone and thresholding at 0.6, which is then upsampled to the image dimension for visualization. The baseline consistently exhibits dispersed activations, causing inaccuracies such as false positives (bicycle in the first image) and missed detections (car in the first image, incomplete train localization in the second, and missed buses in the third and fourth images). In contrast, our method significantly reduces the irrelevant background activations, producing cleaner, focused activations that accurately highlight relevant objects and recover detections missed by the baseline. Overall, these qualitative results illustrate that our method addresses both the pseudo-label noise and feature-space confusion that occurs due to the domain shift, resulting in improved detection accuracy and reliability.


\end{document}